\documentclass[letterpaper, 10 pt, conference]{ieeeconf}  

\IEEEoverridecommandlockouts                              

\usepackage[table,xcdraw]{xcolor}
\usepackage{amssymb}
\usepackage{amsmath}
\usepackage{multirow}
\usepackage{multicol}
\usepackage{graphicx}
\usepackage{subcaption}

\definecolor{rblue}{rgb}{0,0.5,1}
\definecolor{awesome}{rgb}{1.0, 0.13, 0.32}
\definecolor{hollywoodcerise}{rgb}{0.96, 0.0, 0.63}
\definecolor{lasallegreen}{rgb}{0.03, 0.47, 0.19}
\definecolor{hanpurple}{rgb}{0.32, 0.09, 0.98}
\definecolor{green(pigment)}{rgb}{0.0, 0.65, 0.31}
\definecolor{mplblue}{RGB}{31,119,180}
\definecolor{mplorange}{RGB}{255,127,14}
\definecolor{mplgreen}{RGB}{44,160,44}

\definecolor{mygray}{gray}{.9}
\usepackage{booktabs}

\makeatletter
\let\NAT@parse\undefined
\makeatother
\usepackage[pagebackref=false,breaklinks=true,colorlinks,bookmarks=false]{hyperref}
\hypersetup{colorlinks=true,linkcolor={red},citecolor={hanpurple},urlcolor={magenta}}

\title{\LARGE \bf
Not an Obstacle for Dog, but a Hazard for Human:\\A Co-Ego Navigation System for Guide Dog Robots}

\author{Ruiping Liu$^{1,*\dagger}$, Jingqi Zhang$^{1,*}$, Junwei Zheng$^{1,3}$, Yufan Chen$^1$, Peter Seungjune Lee$^1$, Di Wen$^1$,\\ Kunyu Peng$^{1,4}$, Jiaming Zhang$^2$, Kailun Yang$^2$, Katja Mombaur$^1$, and Rainer Stiefelhagen$^1$
\thanks{* indicates equal contribution.}
\thanks{$^1$ The authors are with Karlsruhe Institute of Technology, 76131 Karlsruhe, Germany.}
\thanks{$^2$ The authors are with Hunan University, Changsha 410012, China.
}
\thanks{$^3$ The author is with ETH Zurich, 8092 Zurich, Switzerland.}
\thanks{$^4$ The author is with INSAIT, Sofia University ``St. Kliment Ohridski'', 1784 Sofia, Bulgaria.}
\thanks{$\dagger$ corresponding author. (E-mail: {\tt\small ruiping.liu@kit.edu})}
}
\usepackage{amsmath}
\usepackage{amssymb}
\usepackage{svg}
\usepackage{graphicx}

\begin{document}

\maketitle
\thispagestyle{empty}
\pagestyle{empty}

\begin{abstract}

Guide dogs offer independence to Blind and Low-Vision (BLV) individuals, yet their limited availability leaves the vast majority of BLV users without access. Quadruped robotic guide dogs present a promising alternative, but existing systems rely solely on the robot's ground-level sensors for navigation, overlooking a critical class of hazards: obstacles that are transparent to the robot yet dangerous at human body height, such as bent branches. We term this the viewpoint asymmetry problem and present the first system to explicitly address it. Our Co-Ego system adopts a dual-branch obstacle avoidance framework that integrates the robot-centric ground sensing with the user's elevated egocentric perspective to ensure comprehensive navigation safety. Deployed on a quadruped robot, the system is evaluated in a controlled user study with sighted participants under blindfold across three conditions: unassisted, single-view, and cross-view fusion. Results demonstrate that cross-view fusion significantly reduces collision times and cognitive load, verifying the necessity of viewpoint complementarity for safe robotic guide dog navigation. The source code will be made publicly available at \url{https://github.com/RuipingL/Co-Ego}.

\end{abstract}

\section{Introduction}
Navigating complex environments independently remains a fundamental challenge for people who are Blind or have Low Vision (BLV). While guide dogs have served as trusted mobility aids for decades, their limited availability poses a serious accessibility gap: training a single guide dog costs upward of $\$40,598$~\cite{wirth2008economic}, and about $5\%$ of BLV individuals who could benefit from one have access~\cite{murillo2016cane}. 
Robotic guide dog systems built on quadruped platforms~\cite{hwang2025guidenav, chen2024idog, hwang2023system} offer a promising alternative, combining autonomous locomotion with the physical guidance capabilities of a real guide dog.
\begin{figure}[t]
    \centering
    \includegraphics[width=\linewidth]{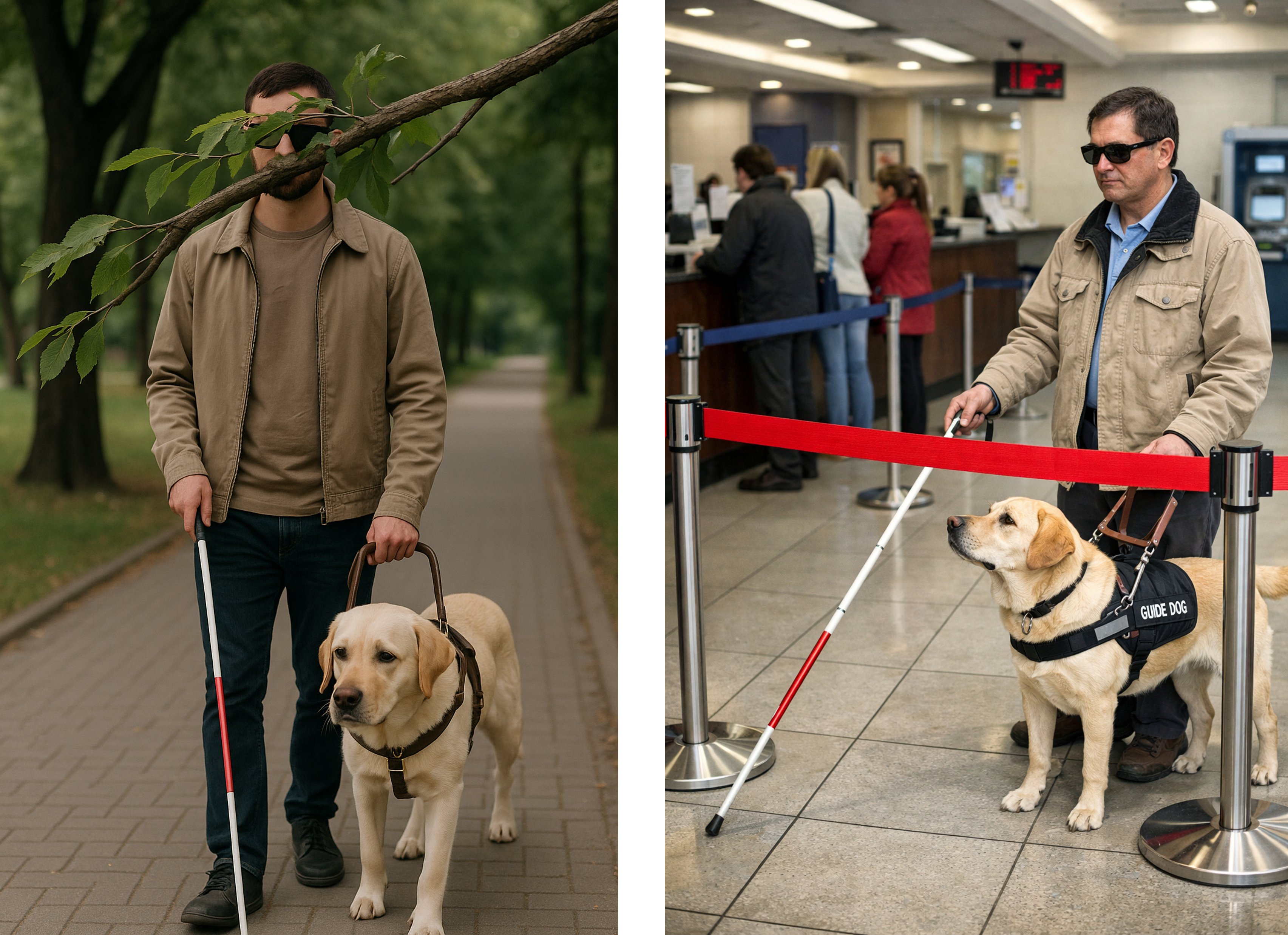}
    \caption{Corner cases where the human encounters hazards that are not obstacles for the guide dog~\cite{hwang2024towards}.}
    \label{fig:corner_cases}
\end{figure}
However, existing robotic navigation systems share a critical blind spot. Most obstacle avoidance methods are designed from the robot’s perspective~\cite{chen2026sea,scheidemann2025obstacle,Falanga20dynamic,forrai2023event}: sensors mounted at platform height, typically 30–50 cm above the ground, detect objects that obstruct the robot’s path, yet overlook hazards that are specifically dangerous to a standing human. 
As discussed in~\cite{hwang2024towards} and illustrated in Fig.~\ref{fig:corner_cases}, examples such as bent branches and queue barriers may remain above the robot’s sensing horizon while still posing substantial collision risks to a BLV user’s head and torso. We define this class as \textit{human-specific obstacles}, \textit{i.e.}, objects that the robot can safely pass under or around, but that nonetheless require a timely warning for the person following behind.

Addressing this gap requires integrating perception from two complementary viewpoints: the robot's egocentric camera, which captures the ground-level scene, and a neck-worn camera positioned at the user's torso height, which captures the user's exposure zone. By leveraging both viewpoints, the proposed Co-Ego system employs a dual-branch architecture that combines robot-centric traversability sensing with human-centered hazard perception for safer shared navigation.

To evaluate the system, we conduct a controlled user study with sighted participants navigating under simulated blindness (blindfold). Participants complete structured routes in environments containing both standard obstacles and human-specific obstacles, under three conditions: unassisted, single-view (robot camera only), and full cross-view fusion (our Co-Ego system). We assess performance through objective metrics (task completion time and collision times) alongside subjective measures including NASA-TLX cognitive load and perceived safety.

The contributions of this work are threefold: (1) We are the first to identify and formalize the viewpoint asymmetry problem in robotic guide dog navigation, where obstacles hazardous to the human user remain undetected by the robot's ground-level sensors. (2) We present a real-time cross-view safety framework for a quadruped guide robot, integrating robot-mounted perception with a user-worn elevated viewpoint to capture hazards overlooked by robot-centric sensing; and (3) we provide empirical evidence from a controlled user study demonstrating that cross-view fusion reduces obstacle collision times and cognitive load during navigation.

\section{Related Work}
\subsection{Robotic Guide Dogs for People with Visual Impairments}










Quadruped robots offer a scalable alternative to guide dogs, which remain scarce due to high costs and lengthy training. 
To better understand how such robots should support blind and vision-impaired users, recent human-centered studies have systematically explored user requirements. 
Kim~\textit{et al.}~\cite{kim2025understanding} identify user preferences for learning-based locomotion and semantic communication, while Hwang~\textit{et al.}~\cite{hwang2023system} develop collaborative indoor navigation based on guide dog handler interviews. 
Doore~\textit{et al.}~\cite{doore2024co_design} demonstrate that industrial quadrupeds like Boston Dynamics Spot can reproduce simple navigation tasks, whereas Wang~\textit{et al.}~\cite{wang2023can} find that walking noise disrupts the acoustic spatial awareness relied upon by BLV users, highlighting the importance of aligning robot behavior with user perception. 
However, these studies primarily rely on Wizard-of-Oz evaluations rather than deployment on autonomous robotic guide dogs.


Practical strategies such as flexible leash connections~\cite{xiao2021robotic}, comfort-based traction adjustment~\cite{chen2023quadruped_guidance}, and reinforcement learning approaches robust to human tugging forces~\cite{defazio2023seeing_quadruped} have demonstrated improvements in robot locomotion stability during physical Human-Robot Interaction (pHRI). 
At the perception and system level, Han~\textit{et al.}~\cite{han2025space} propose space-aware instruction tuning for environmental understanding, while comprehensive platforms such as RDog~\cite{cai2024navigating} integrate mapping, force feedback, and voice cues, DogSurf~\cite{bazhenov2024dogsurf} addresses terrain safety through surface recognition, and iDog~\cite{chen2024idog} augments biological guide dogs with AI-assisted obstacle detection. While Chen~\textit{et al.}~\cite{chen2023quadruped_guidance} account for the user's inscribing radius in collision avoidance, the current robotic guide dog research remains primarily focused on the robot's mobility and obstacle awareness, with limited attention to supporting shared mobility between the robot and the user.


\subsection{Traversability Perception and Hazard Negotiation for People with Visual Impairments}





Safe navigation requires accurate perception of walkable surfaces, obstacles, and environmental hazards. 
Recent wearable systems address this through multimodal sensing~\cite{gao2025wearable_obstacle_avoidance}.
eLabrador~\cite{kan2025elabrador} integrates RGB-D cameras with GPS for outdoor guidance via audio-haptic feedback. 
For scene understanding, He~\textit{et al.}~\cite{he2026parameter_3D_ssc} propose adaptive attention for monocular 3D semantic scene completion on embedded hardware. Trans4Trans~\cite{zhang2022trans4trans} enables transparent obstacle segmentation to detect glass doors and walls. 
Son~\textit{et al.}~\cite{son2025infrastructure} leverage prior maps and sensing nodes for infrastructure-enabled dynamic path adjustment. 
WalkGPT~\cite{sultan2026walkgpt} introduces a pixel-grounded vision-language model for depth-aware accessibility guidance. 
ObjectFinder~\cite{liu2024objectfinder} supports open-vocabulary object search in unfamiliar environments. Some wearable assistive systems consider head-level obstacles~\cite{klimesova2024head, munoz2025embedded, Santos2025nav}, whereas our work is the first to integrate this capability with the mobility support of a robotic guide dog.

Several datasets also advance this field. SideGuide~\cite{park2020sideguide} provides large-scale sidewalk annotations for accessible path planning, while TBRSD~\cite{chen2023atmospheric} addresses blind road segmentation under varying atmospheric and thermal conditions. mmWalk~\cite{ying2025mmwalk}, SANPO~\cite{Waghmare2025sanpo}, and GuideDog~\cite{kim2025guidedog} provide multi-view egocentric annotations for outdoor navigation. Situat3DChange~\cite{liu2025situated} enables understanding of environmental changes in familiar spaces.

\subsection{Navigation for Quadruped Robots}
Recent advances in quadruped navigation span semantic understanding, robust locomotion, and safety~\cite{joventino2025multi_agent,you2026smoothturn,tang2026zero_velocity,zhu2026trans,li2026task_level_gait,zhang2025embodied_navigation,wang2025x_nav,elnoor2024amco}. 
For semantic navigation, VLFM~\cite{yokoyama2024vlfm} enables zero-shot goal-reaching via vision-language features, NaVILA~\cite{cheng2024navila} and StreamVLN~\cite{wei2025streamvln} address vision-and-language navigation with hierarchical control and streaming context, and Becoy~\textit{et al.}~\cite{becoy2025autonomous} demonstrate coverage planning using topological maps.
For robust locomotion, Gao and Awuah-Offei~\cite{gao2025navigation_mine,gao2026efficient_autonomous} and Aditya~\textit{et al.}~\cite{aditya2025robust} tackle challenging environments including underground mines, whereas Kim~\textit{et al.}~\cite{kim2025high_speed_control_navigation} and Hoeller~\textit{et al.}~\cite{hoeller2024anymal} achieve agile traversal on complex terrain. 
For safety, SEA-Nav~\cite{chen2026sea} uses differentiable control barrier functions, and Lin~\textit{et al.}~\cite{lin2025one_filter} propose a unified filter for unknown environments.
In spite of these advances, existing methods focus on robot-centric mobility rather than human-centered guidance for visually impaired users.



\section{Method}
In this section, we detail the Co-Ego system. To address the inherent limitations of single-view perception in dynamic environments, we design a dual-branch architecture. This architecture leverages high-resolution onboard depth sensing on the robot for local path planning and couples it with an asynchronous smartphone-based perception branch for auxiliary collision avoidance.

\subsection{Hardware}
\begin{figure}
    \centering
    \includegraphics[width=0.55\linewidth]{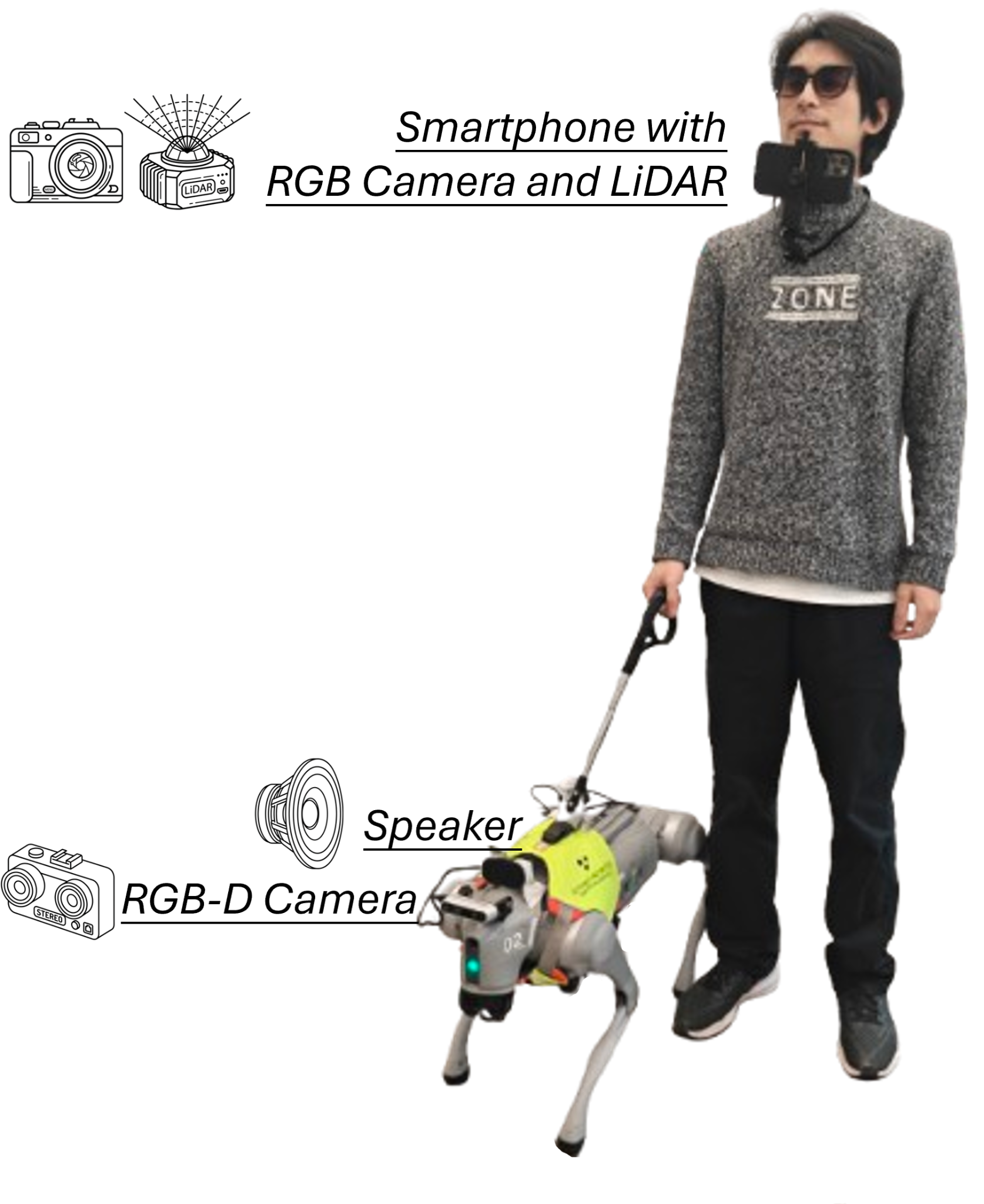}
    \caption{Hardware components of our Co-Ego system.}
    \label{fig:hardware}
\end{figure}

As illustrated in Fig.~\ref{fig:hardware}, our system consists of a quadrupedal robot integrated with a wearable perception module centered on humans, seamlessly combining robotic mobility with user-centered perception.

\subsubsection{Quadrupedal Robot Platform}
The primary robotic agent is a Unitree Go2 quadruped, selected for its dynamic stability and robust onboard computing capabilities. To facilitate local path planning, a front-mounted Intel RealSense D435i depth camera provides high-fidelity spatial data. 
Furthermore, an external speaker is integrated to issue dynamic obstacle warnings that enhance the shared situational awareness between the user and the autonomous system.

\subsubsection{Human-Centered Wearable Perception}
To mitigate the blind spots inherent to the robot's low-elevation viewpoint, a smartphone equipped with a LiDAR sensor is mounted on the user's chest via a stabilized harness. Acting as an elevated, egocentric perception node, this device drives the asynchronous reactive collision avoidance branch. By leveraging its integrated depth sensing, the wearable node effectively detects overhanging or elevated hazards that are typically occluded from the robot's onboard sensors.

\subsection{Obstacle Avoidance: Dual-Perception Priority-Based Navigation}
\label{sec:methodology_method1}

\begin{figure*}
    \centering
    \includegraphics[width=\textwidth]{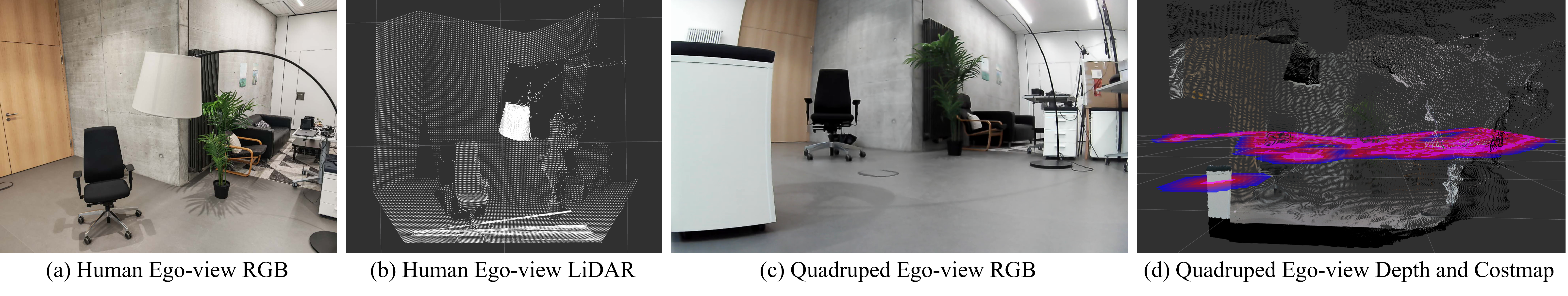}
    \caption{Viewpoint asymmetry between human and robot egocentric views. (a) RGB and (b) point cloud from the user's neck-worn camera capture the hanging lamp at head height but miss the ground-level cabinet. (c) RGB and (d) cost map from the quadruped's point cloud detect ground-level obstacles (red regions) with a navigable path in between (blue regions) but remain blind to the overhead lamp.}
    \label{fig:visualization}
\end{figure*}

\subsubsection{System Kinematics and Coordinate Frames}

To formalize the system's spatial configuration, we define the robot base frame $\mathcal{F}_R$, alongside the physical ($\mathcal{F}_{P,k}$) and optical ($\mathcal{F}_{O,k}$) frames for both the egocentric phone camera ($k=human$) and the robot-mounted depth camera ($k=dog$). 

The mapping from each optical frame to the physical robot base is formulated as a two-stage transformation in $SE(3)$:
\begin{equation}
    \mathbf{T}_{O,k}^R(t) = \mathbf{T}_{P,k}^R(t) \mathbf{T}_{O,k}^{P,k}, \quad k \in \{human, dog\},
\end{equation}
where $\mathbf{T}_{O,k}^{P,k}$ represents the fixed optical-to-physical transformation. This follows the standard convention where the Z-axis of optical frames points forward, while the X-axis of physical frames aligns with the sensor's heading.

The mounting transformation $\mathbf{T}_{P,k}^R(t)$ is hardware-dependent:
The mounting transformation $\mathbf{T}_{P,k}^R(t)$ is hardware-dependent. Specifically, while the robot-side camera ($k=\text{dog}$) is rigidly affixed to the chassis---rendering $\mathbf{T}_{P,\text{dog}}^R$ a constant extrinsic matrix---the phone-side camera ($k=\text{human}$) is worn by the user. Consequently, its pose $\mathbf{T}_{P,\text{human}}^R(t)$ is highly dynamic and must be continuously estimated to capture the stochastic human-robot relative motion and ensure spatial synchronization.

\subsubsection{Local Mapping and APF-Based Navigation}

At each time step, the robot-mounted camera captures a depth image, which is projected into a 3D point cloud and transformed to the robot base frame $\mathcal{F}_R$ via the fixed extrinsic matrix $\mathbf{T}_{O,dog}^R$. We apply a $Z$-axis pass-through filter to eliminate ground plane reflections and overhead structures beyond the robot's physical clearance height. The remaining points are orthogonally projected onto the $XY$-plane to construct a 2D local costmap, where occupied cells are inflated by a radius $R_{inf}$ to explicitly account for the robot's physical footprint.

Using the continuously updated local costmap, an Artificial Potential Field (APF) algorithm generates the baseline navigation commands. The workspace is modeled as a scalar field combining an attractive potential towards the 2D goal $\mathbf{p}_{goal}$ and a repulsive potential pushing the robot away from the inflated obstacles. The resulting desired force vector, derived from the negative gradient of these combined potential fields, is denoted as $\mathbf{F}_{APF} = [F_x, F_y]^T$. Finally, this planar force vector is mapped through an admittance controller to yield the baseline velocity command $\mathbf{v}_{APF} = [v_{x, APF}, v_{y, APF}]^T$.

\begin{figure}
    \centering
    \includegraphics[width=\linewidth]{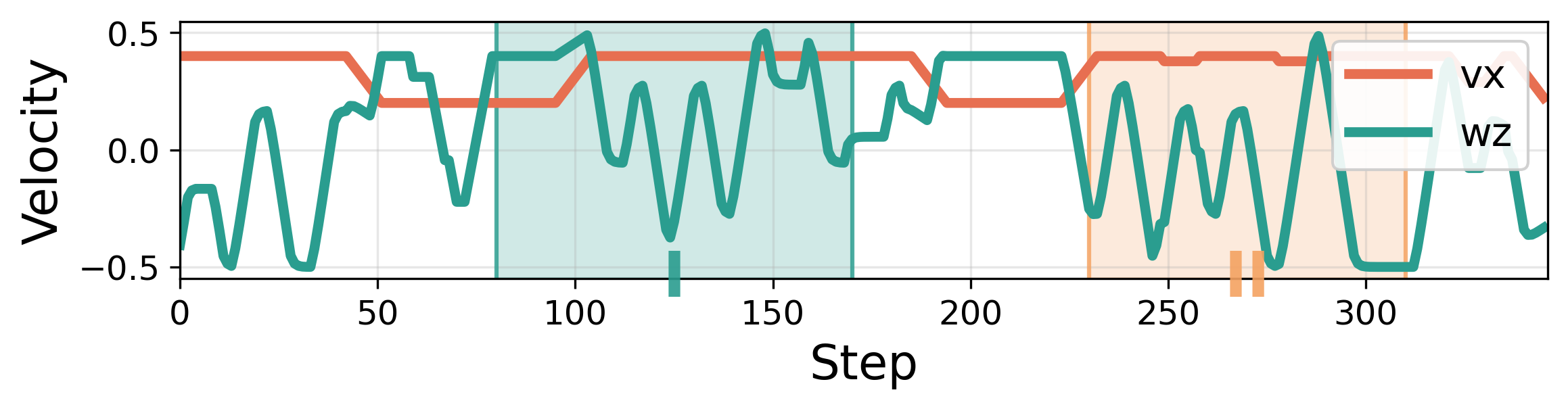}
    \caption{Change of velocity and turns when avoiding an obstacle. $v_x$ denotes the forward velocity, while $w_z$ denotes the angular velocity. Zone I corresponds to turning to avoid the obstacle, whereas Zone II corresponds to recovery.}
    \vspace{-2mm}
    \label{fig:velocity}
\end{figure}

\subsubsection{Smartphone-Assisted Reactive Perception}
While APF-based navigation provides mathematically smooth local trajectories for the quadruped, the robot’s onboard perception is fundamentally constrained by its low-elevation viewpoint, which is insufficient for safeguarding the human user in complex environments. Specifically, the robot-side sensors focus on ground-level traversability and immediate obstacles within its own footprint, often failing to account for overhanging hazards or  dynamic entities that pose a direct risk to the human. 
Fig.~\ref{fig:visualization} compares the two views, where a,b represent the human view and c,d denote the robot view.

To bridge this gap and ensure comprehensive navigation safety, a smartphone is integrated as an independent, elevated perception node. By positioning the phone at the user’s chest level, we introduce an egocentric viewpoint that complements the robot’s local sensing. This dual-branch architecture allows the system to not only manage the robot's locomotion but also proactively avoid obstacles specifically relative to the human’s height and trajectory, providing a crucial layer of safety redundancy that a standalone robot cannot achieve.

Since robot-centric obstacle avoidance alone cannot ensure human-level safety, we introduce an elevated wearable sensing branch, where the smartphone’s depth sensor captures supplementary depth maps in its optical frame $\mathcal{F}_{O,ego}$ at a high frame rate to detect human-specific hazards and trigger reactive avoidance when needed.  Rather than building a dense local map, it directly uses the raw point cloud data to detect nearby obstacles and estimate their relative position to the human. When an obstacle falls within a critical safety margin, the smartphone bypasses the APF planner and generates a reactive evasive or braking command, $\mathbf{v}_{human}$, which is executed with higher priority than the baseline $\mathbf{v}_{APF}$ generated by the dog. This enables rapid collision avoidance even if the primary command pipeline is compromised.

\subsubsection{Asynchronous Multiplexing and Priority Hierarchy}
\begin{figure}
    \centering
    \includegraphics[width=0.7\linewidth]{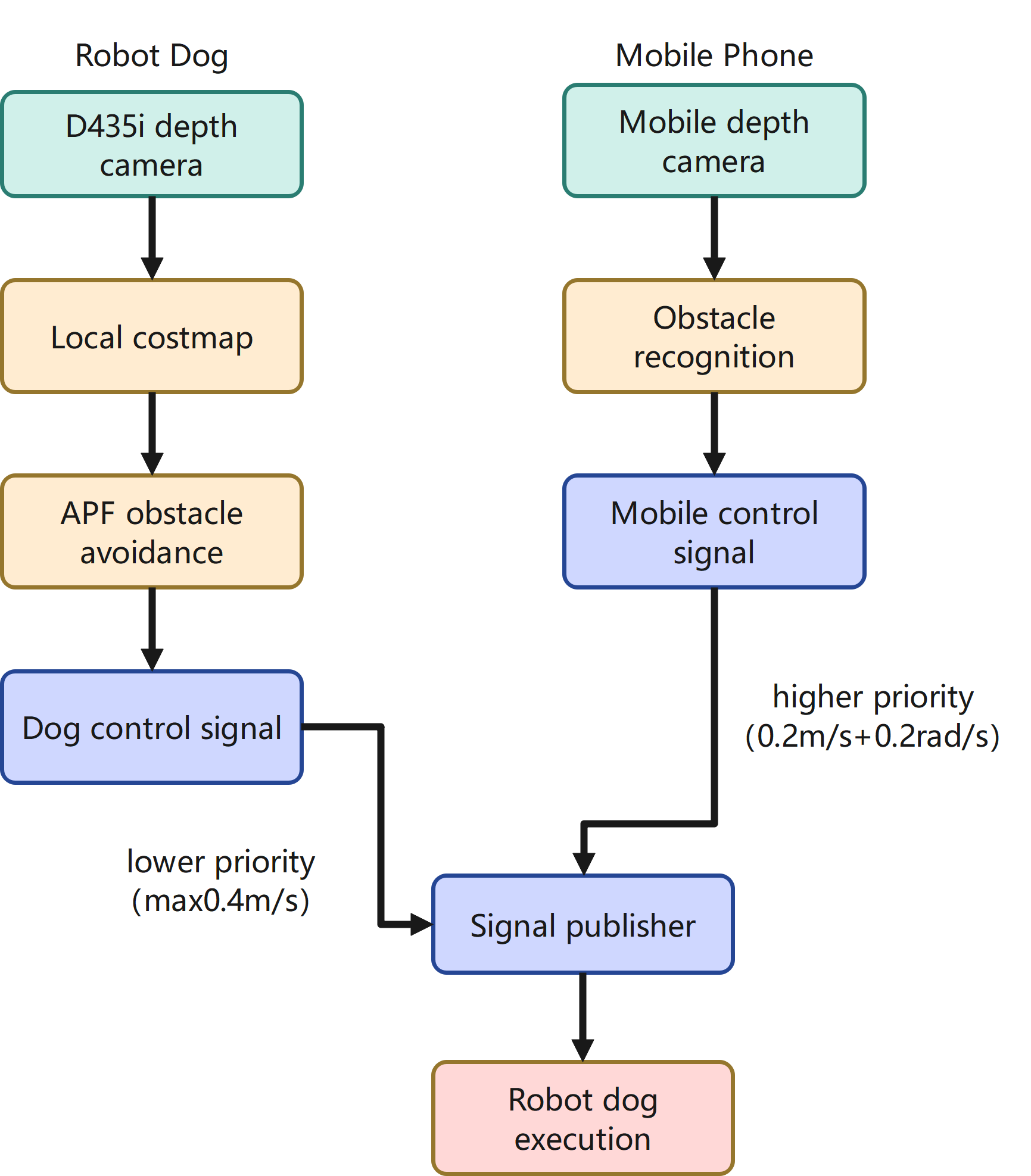}
    \caption{Flowchart of the dual-perception priority-based navigation algorithm.}
    \label{fig:method1}
\end{figure}

The final crucial component of this method is the signal fusion module (Fig.~\ref{fig:method1}). Both the onboard compute unit (publishing $\mathbf{v}_{APF}$) and the smartphone (publishing $\mathbf{v}_{human}$) operate asynchronously. They publish their respective command vectors to a central topic managed by the robot's low-level locomotion controller.

To prevent conflicting commands that could lead to kinematic singularity or collision, we propose a strict priority-based arbitration mechanism. The smartphone acts as a high-priority emergency override. The executed velocity command $\mathbf{v}_{cmd}(t)$ is governed by a boolean arbitration function $A(t)$:
\begin{equation}
    A(t) = 
    \begin{cases}
        1, & \text{if } \|\mathbf{v}_{human}(t)\| > \epsilon \\
        0, & \text{otherwise}
    \end{cases},
\end{equation}
where $\|\mathbf{v}_{human}(t)\|$ represents the magnitude of the received human control signal, and $\epsilon$ is a predefined deadband threshold used to filter out joystick drift or baseline noise. The final velocity command is computed via the arbiter as follows:
\begin{equation}
    \mathbf{v}_{cmd}(t) = A(t) \cdot \mathbf{v}_{human}(t) + (1 - A(t)) \cdot \mathbf{v}_{APF}(t),
\end{equation}

This formulation guarantees that, whenever the smartphone detects an imminent frontal hazard (\(A(t)=1\)), its control command takes priority over the default APF output as shown in Fig.~\ref{fig:velocity}. As a result, the quadruped follows the APF for smooth and stable navigation under normal conditions, while the smartphone can immediately override it with an emergency command when an obstacle is detected.

\subsection{Speech Interaction in Navigation}
\label{sec:methodology_semantic_hri}

Beyond low-level geometric obstacle avoidance, assistive robotics also requires high-level scene understanding and effective communication. This is especially critical for visually impaired users, who lack direct visual situational awareness and therefore depend heavily on environmental context to establish cognitive trust in the system. We implement an audio interaction module, deploying an edge-optimized Vision-Language Model (VLM) directly onboard the quadruped robot.
\begin{figure}
    \centering
    \includegraphics[width=0.7\linewidth]{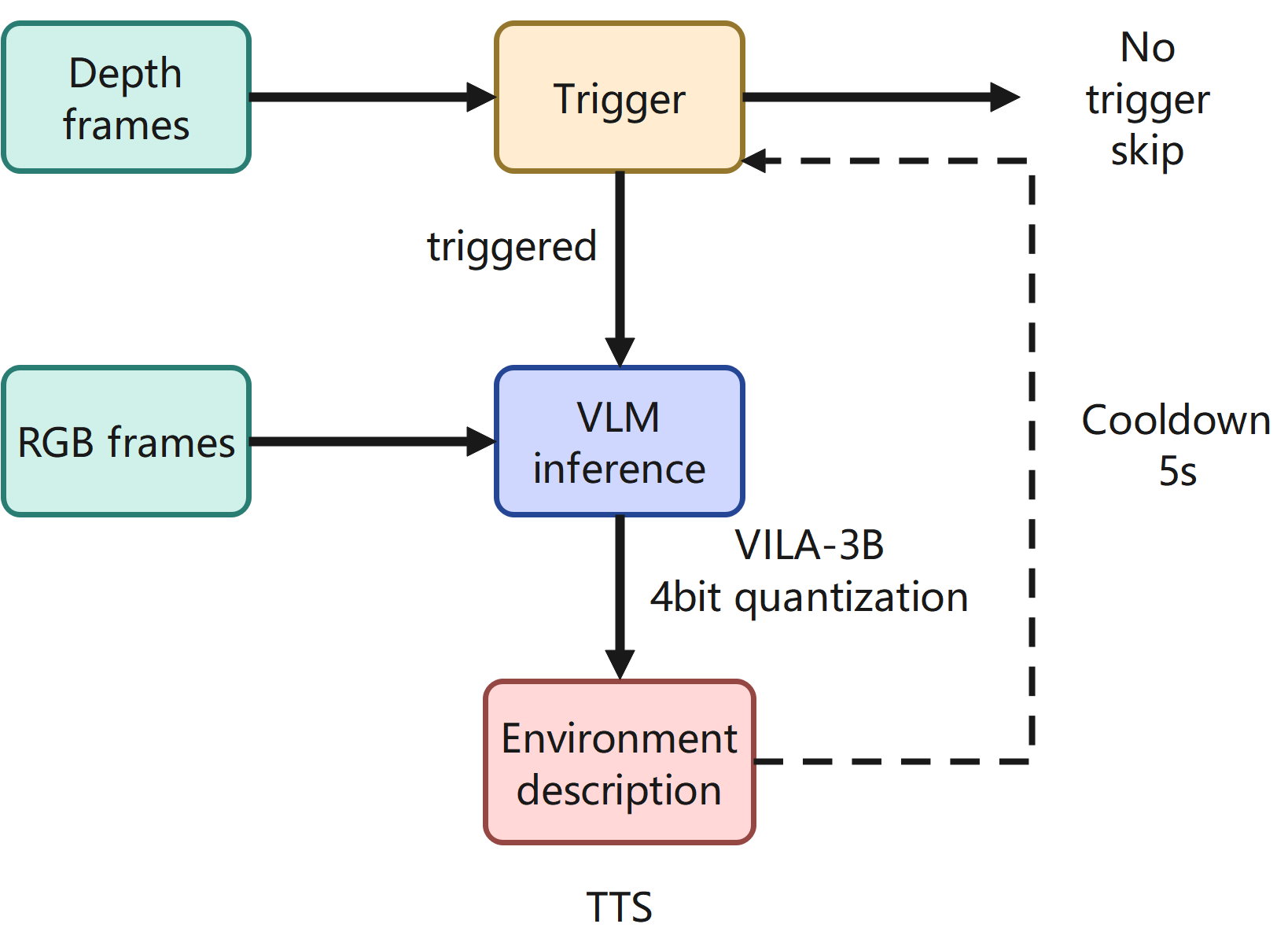}
    \caption{Flowchart of the environment warning process.}
    \label{fig:method2}
\end{figure}
Continuously processing a video stream through the VLM is computationally prohibitive on edge devices and may also lead to auditory fatigue in assistive settings. Furthermore, since the selected VLM solely relies on RGB inputs, it inherently lacks depth perception. To bridge this gap, we design an event-triggered, decoupled architecture: \textit{``Geometric-Trigger, Semantic-Inference.''} as shown in Fig.~\ref{fig:method2}.

Instead of relying on a dense global costmap, we utilize direct depth map thresholding within a defined Region of Interest (ROI) as a lightweight geometric sentinel. Given a continuous depth stream, let the depth map denote the depth map captured at a specific time. We define an ROI on the image plane corresponding to the user's immediate forward traversal path. Let the minimum depth value represent the minimum valid depth value (\textit{i.e.}, the distance to the nearest obstacle) extracted from the pixels within the ROI, and let the critical depth be a predefined critical safety threshold.

When the hazard condition is met, the system captures a synchronized RGB frame from robot's view and feeds it to the edge-native model together with a hazard-specific prompt, (\textit{i.e.}, \textit{``An obstacle is very close. Briefly describe what it is and its relative position.''}) The model then generates a concise semantic output sequence describing the threat. This description is finally delivered to the visually impaired user through an integrated Text-to-Speech (TTS) interface, thereby completing the assistive perception loop.

\section{Experiments}
\begin{figure}
    \centering
    \includegraphics[width=0.5\linewidth]{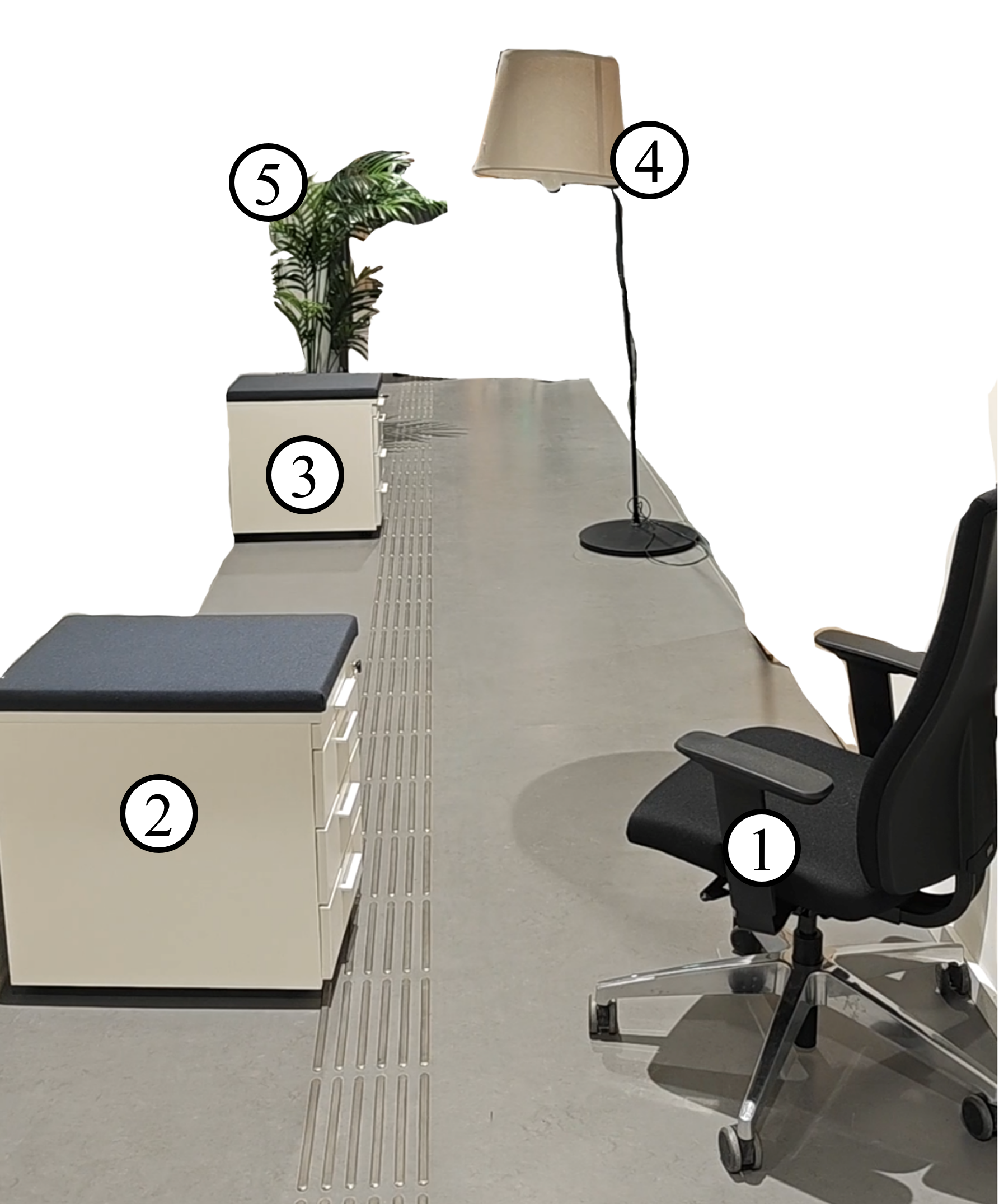}
    \caption{Test scenario. The hallway includes three ground-level obstacles affecting both the robot dog and the human (a chair and two cabinets), and two hanging obstacles affecting only the human (a lamp and a bent branch).}
    \label{fig:test_scenario}
\end{figure}

\subsection{Participants and Procedure}
\begin{figure}
    \centering
    \includegraphics[width=\linewidth]{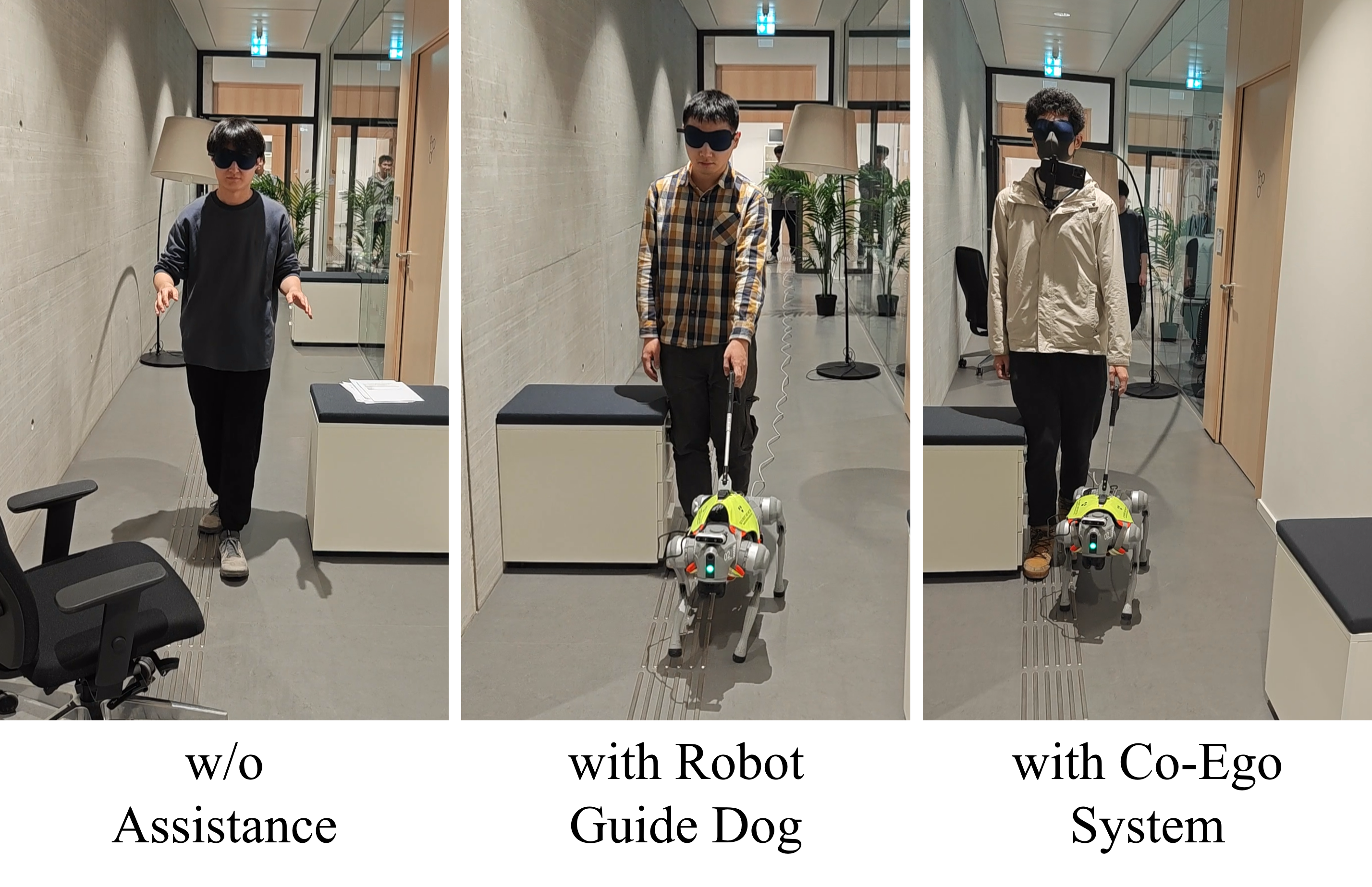}
    \caption{Illustration of the three navigation conditions evaluated in our study.}
    \label{fig:baselines}
    \vspace{-2mm}
\end{figure}

We recruited six sighted participants to navigate the hallway while blindfolded and avoid five obstacles, including three obstacles that obstructed both the robot dog and the human (two cabinets and one chair) and two obstacles that posed hazards only to the human (a bent branch and a lamp), as shown in Fig.~\ref{fig:test_scenario}. We report participants’ heights, as height may influence walking pace and thus the experience with the system: five male participants measured $190$ cm, $183$ cm, $183$ cm, $175$ cm, and $173$ cm, and one female participant measured $175$ cm. Each participant completed the route under three conditions: tactile exploration without assistance, with the robot dog, and with our co-ego system (Fig.~\ref{fig:baselines}). The obstacle arrangement was randomized for each trial. We recorded task completion time, the number of collisions, perceived safety on a Likert scale, and cognitive load using NASA-TLX. In addition to these quantitative measures, we conducted semi-structured interviews with questions such as \textit{``Was there a moment when you felt unsafe?''} and asked participants about their preferences across the three conditions.

\subsection{Results}
\begin{figure*}[t]
    \centering
    \begin{subfigure}[t]{0.31\textwidth}
        \centering
        \includegraphics[width=\linewidth]{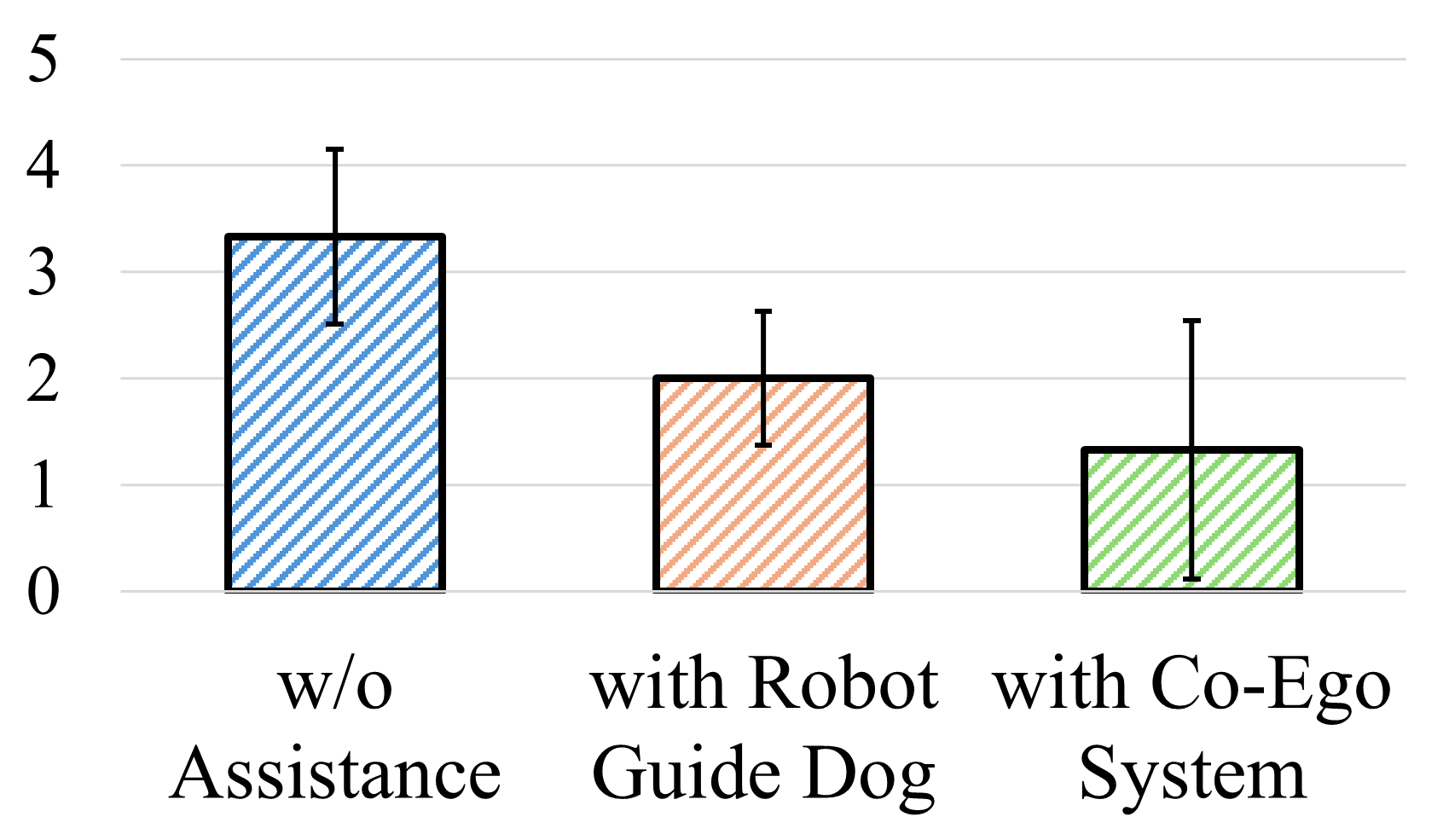}
        \caption{Number of collisions.}
        \label{fig:collision}
    \end{subfigure}
    \hfill
    \begin{subfigure}[t]{0.31\textwidth}
        \centering
        \includegraphics[width=\linewidth]{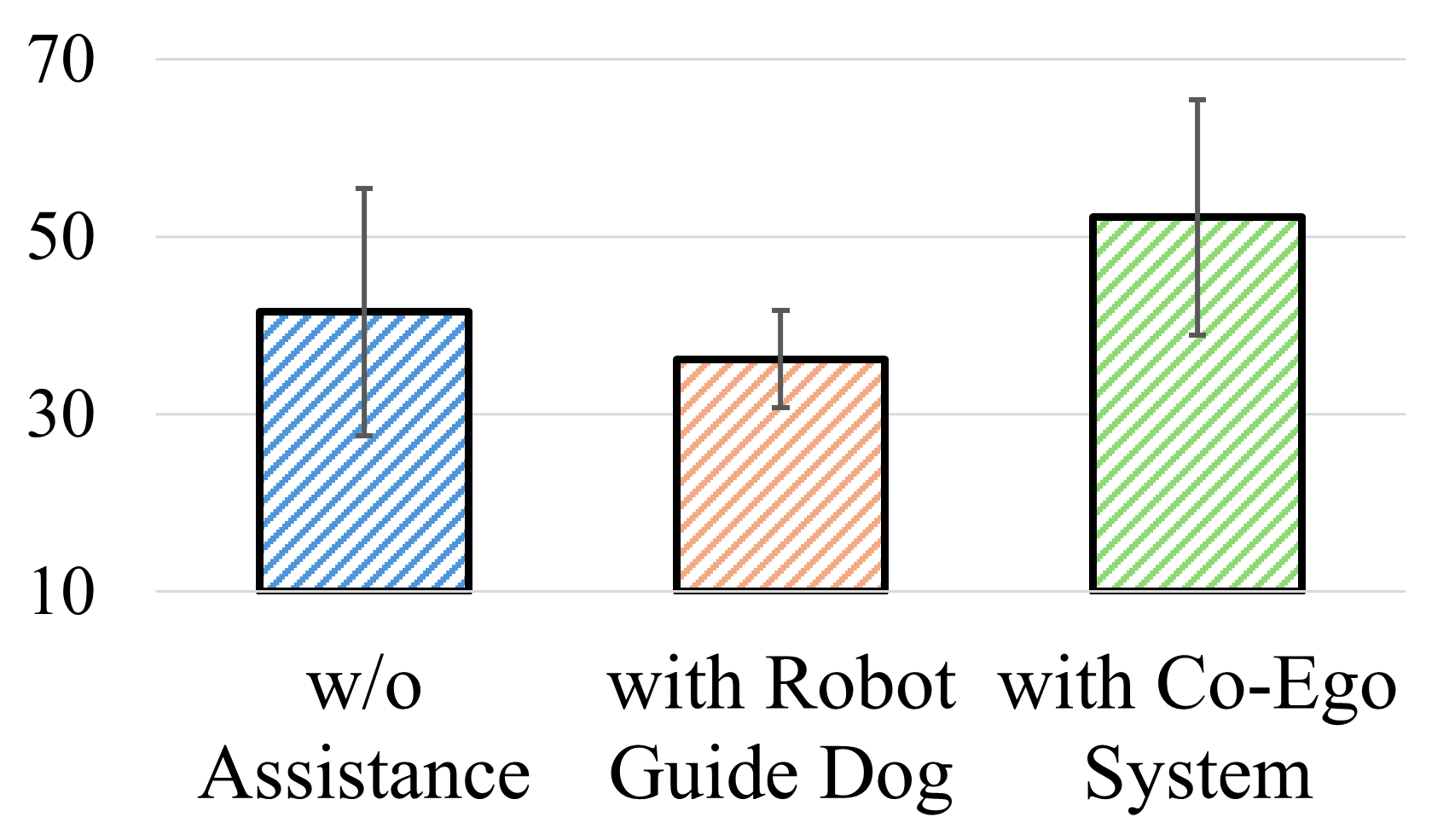}
        \caption{Task completion time.}
        \label{fig:time}
    \end{subfigure}
    \hfill
    \begin{subfigure}[t]{0.31\textwidth}
        \centering
        \includegraphics[width=\linewidth]{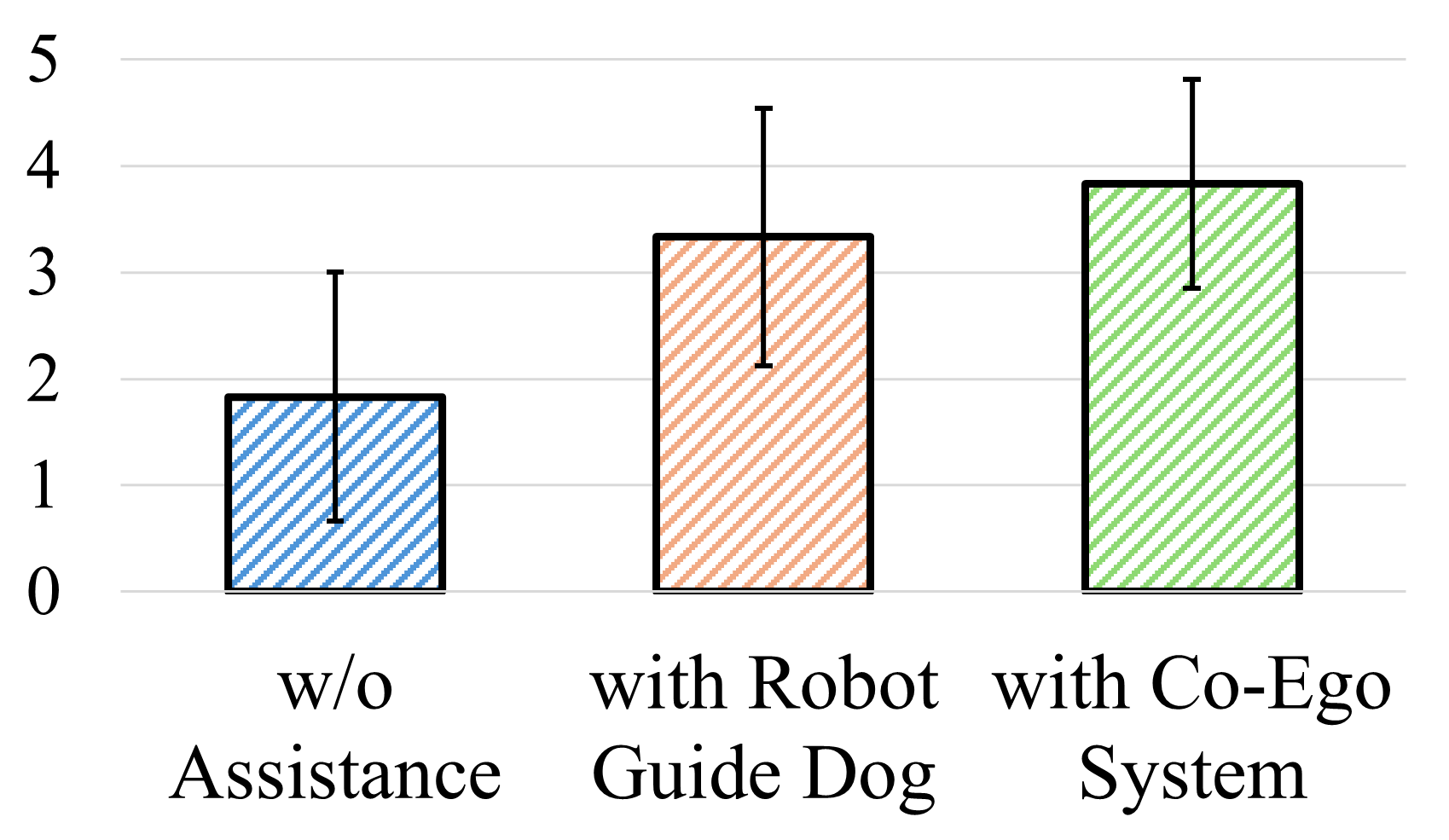}
        \caption{Sense of safety.}
        \label{fig:safety}
    \end{subfigure}
    \caption{Quantitative comparison of the three navigation conditions: without assistance, with robot guide dog, and with our Co-Ego system. From left to right, the plots show (a) number of collisions, (b) task completion time, and (c) perceived safety. Error bars denote participant variability.}
    \vspace{-2mm}
    \label{fig:quantitative}
\end{figure*}

\begin{figure}
    \centering
    \includegraphics[width=\linewidth]{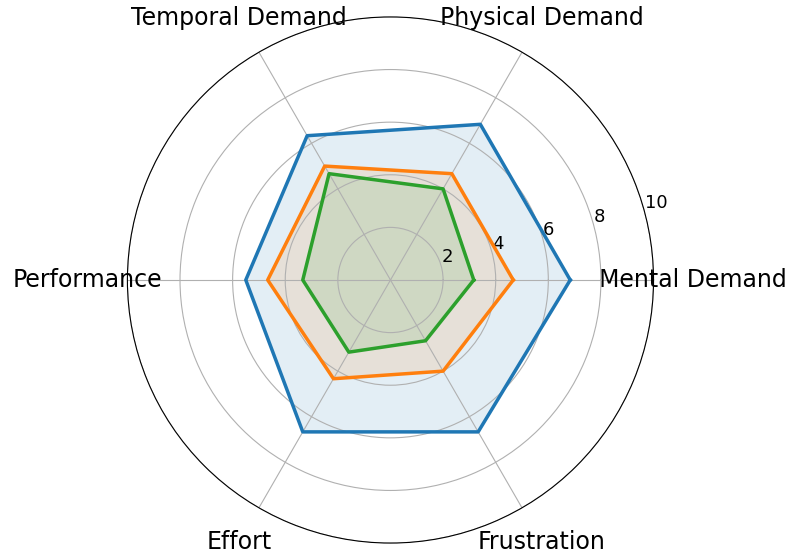}
    \caption{NASA-TLX cognitive load measured while completing tasks under three conditions: \textcolor{mplblue}{without assistance}, \textcolor{mplorange}{with the robot guide dog}, and \textcolor{mplgreen}{with our Co-Ego system}. Lower scores indicate lower cognitive load.}
    \label{fig:nasa_tlx}
    \vspace{-2mm}
\end{figure}

\paragraph{System Performance for Navigation} 
In Fig.~\ref{fig:collision}, we observe that the systems substantially reduced the number of collisions, from $3.33 \pm 0.82$ without assistance to $2.00 \pm 0.63$ with only the robot guide dog, and further to $1.33 \pm 1.21$ with our Co-Ego system. Although the robot guide dog could robustly avoid obstacles on the ground, it did not explicitly account for the participant’s body pose, which still led participants to collide with obstacles. The user's egocentric view compensated for this limitation to a large extent, but it remained vulnerable when participants bent down to follow the dog more closely. In such cases, the camera was directed toward the ground and could no longer capture hanging obstacles. Fig.~\ref{fig:time} shows the task completion time. Using only the robot guide dog ($36.20 \pm 5.51$ s) reduced task completion time relative to the no-assistance condition ($41.51 \pm 13.94$ s). In contrast, our Co-Ego system led to a longer completion time ($52.18 \pm 13.23$ s), likely because participants moved more cautiously around obstacles and paused more frequently to interpret hazards ahead.
\paragraph{Cognitive Load and Sense of Safety}
We used NASA-TLX to evaluate cognitive load during task completion across six dimensions: Temporal Demand, Physical Demand, Mental Demand, Frustration, Effort, and Performance. As shown in Fig.~\ref{fig:nasa_tlx}, pure tactile exploration resulted in the highest cognitive load across all dimensions, followed by the robot guide dog condition, while our Co-Ego system achieved the lowest cognitive load in all dimensions. 
It was surprising that our Co-Ego system required the longest time to complete the task, yet achieved the lowest Temporal Demand. One participant attributed this to \textit{the uncertainty and lack of safety during tactile exploration}, while another noted that \textit{although the Co-Ego system occasionally stopped, they understood that it was detecting and assessing something ahead}. The safer participants felt, the lower the perceived demand was.

As shown in Fig.~\ref{fig:safety}, participants felt safest with our Co-Ego system, followed by the robot guide dog, and least safe during tactile exploration alone. Under tactile exploration alone, participants reported \textit{feeling particularly unsafe when there was nothing nearby to touch}. One participant noted that this was especially the case \textit{when obstacles were lower than the upper body}. Several participants attributed their preference for the systems to their \textit{trust in the system’s ability to process more information} and the \textit{better orientation} they provided. At the same time, some still described situations in which relying passively on the system felt unsafe, as \textit{collisions could be more unsettling when they were being guided rather than acting independently}. In addition, several participants occasionally \textit{bumped into the dog} because they were not yet used to walking alongside it.

\paragraph{Condition Preference}
All participants ranked tactile exploration without assistance as the least preferred option, while five preferred our Co-Ego system over the robot guide dog alone. One participant explained that \textit{the Co-Ego system was the best choice when obstacles were present and the aisle was not too narrow, since otherwise it required too many turns. The robot guide dog alone also provided a strong sense of safety, and could be further improved with adjustable speed. For short paths, the participant preferred tactile exploration.} The only participant who preferred the robot guide dog alone over the Co-Ego system attributed this to \textit{the need to wear an additional device and discomfort with the changing speed}.
\paragraph{Directions for Improvement}
Our robot guide dog currently uses a leash similar to that of a biological guide dog, whereas participants suggested a \textit{rigid connection between the handle and the robot dog}. Some participants also noted that they could not follow the robot at their natural walking pace and therefore occasionally bumped into it. They suggested \textit{enabling velocity adjustment through the handle interaction}, \textit{e.g.}, by pulling handle to slow the robot down.

\section{Limitations and Future Work}
Despite the cross-view perception, inherent blind spots persist due to the limited horizontal field of view of the forward-facing cameras, which could be mitigated by integrating the onboard LiDAR for 360° coverage. Additionally, the current navigation module follows costmap gradients, which is effective for structured paths but lacks flexibility in unstructured environments. Incorporating a VLN module would enable instruction-driven, goal-conditioned traversal while preserving safety constraints. 
In our study setting, collisions with obstacles could not be fully eliminated. Therefore, for safety reasons, we recruited sighted participants wearing blindfolds to allow better control of potential risks. This work represents a first step toward building a Co-Ego system that can help users avoid obstacles at all heights during navigation. In future work, we will further optimize the system based on participant feedback and conduct a large-scale user study with people with visual impairments.

\section{Conclusion}
In this work, we have identified the viewpoint asymmetry problem in robotic guide dog navigation and presented a Co-Ego system with a cross-view obstacle avoidance framework fusing data from the robot's egocentric camera and a user-worn neck camera. A controlled user study under simulated blindness showed that cross-view fusion reduces near-miss events and cognitive load over both an unassisted baseline and a single-view system. These results demonstrate that modeling the user's body-height exposure zone is necessary for safe robotic guide dog navigation. Future work will extend the evaluation to real BLV participants and outdoor environments.

\bibliographystyle{IEEEtran}
\bibliography{main}

\end{document}